\documentclass{article}

\usepackage{arxiv}

\usepackage[utf8]{inputenc} 
\usepackage[T1]{fontenc}    
\usepackage{hyperref}       
\usepackage{url}            
\usepackage{booktabs}       
\usepackage{amsfonts}       
\usepackage{nicefrac}       
\usepackage{microtype}      
\usepackage{lipsum}
\usepackage{graphicx}
\usepackage{diagbox}
\usepackage{slashbox}
\usepackage{subfig}

\graphicspath{ {./images/} }

\title{Synthetic Demographic Data Generation for Card Fraud Detection Using GANs}

\author{
 Shuo Wang \\
  Department of Computer Science\\
  Memorial University of Newfoundland\\
  St. John's, NL A1C5S7 \\
  \texttt{shuow@mun.ca} \\
   \And
 Terrence Tricco \\
  Department of Computer Science\\
  Memorial University of Newfoundland\\
  St. John's, NL A1C5S7 \\
  \texttt{tstricco@mun.ca} \\
  \And
 Xianta Jiang \\
  Department of Computer Science\\
  Memorial University of Newfoundland\\
  St. John's, NL A1C5S7 \\
  \texttt{xiantaj@mun.ca} \\
  \And
 Charles Robertson \\
  Verafin\\
  St. John's, NL A1A 0L9 \\
  \texttt{charles.robertson@verafin.com} \\
  \And
 John Hawkin \\
  Verafin\\
  St. John's, NL A1A 0L9 \\
  \texttt{John.Hawkin@verafin.com} \\
}

\begin{document}
\maketitle
\begin{abstract}
Using machine learning models to generate synthetic data has become common in many fields. Technology to generate synthetic transactions that can be used to detect fraud is also growing fast. Generally, this synthetic data contains only information about the transaction, such as the time, place, and amount of money. It does not usually contain the individual user's characteristics (age and gender are occasionally included). Using relatively complex synthetic demographic data may improve the complexity of transaction data features, thus improving the fraud detection performance. Benefiting from developments of machine learning, some deep learning models have potential to perform better than other well-established synthetic data generation methods, such as microsimulation. In this study, we built a deep-learning Generative Adversarial Network (GAN), called DGGAN\footnote{Our DGGAN model is open-sourced at https://github.com/MountStonne/DGGAN}, which will be used for demographic data generation. Our model generates samples during model training, which we found important to overcame class imbalance issues. This study can help improve the cognition of synthetic data and further explore the application of synthetic data generation in card fraud detection.\\
\end{abstract}


\section{Introduction}
With the increasing number of online transactions, card fraud detection is indispensable for business activities.
One of the commonly used fraud detection approach is to label existing normal and fraudulent transactions and identify suspected fraud cases by learning the differences. 
However, applying effective card fraud identification methods in existing commercial activities is very difficult. This is mainly due to the inability to obtain real fraud cases efficiently since both the law and most corporate policies prioritize protecting user privacy as much as possible.
Even with access to real data, there is often insufficient data to train fraud detection models. Therefore, synthetic transaction generation technology has become key for improving fraud detection.

Existing synthetic transaction generators, such as Banksformer \cite{Nickerson2022Banksformer} and PaySim \cite{lopez2016paysim}, only generate transaction information. They do not generate rich information about the individuals that execute these transactions. An individual's transactions are significantly impacted by their demographics, including age, gender, occupation, and more. While previous approaches have been effective in enhancing the performance of fraud detection, incorporating individual-level demographic data could further improve its accuracy.


In this study, we focus on synthetic generation of rich demographic data using a Generative Adversarial Network (GAN) \cite{goodfellow_generative_2020}. To the best of our knowledge, this is the first research related to generating demographic data using GANs for card fraud detection. After generation using this deep learning method,  we visualize and evaluate the synthetic data using various metrics. The results of this process will lead to new insights into demographic data generation. Coupling high-quality synthetic demographic data will lead to better generation of financial transaction data.

\section{Methods}
\label{sec:headings}
\subsection{Dataset}
Considering the inaccessibility of demographic data in this area, we use two public data sets. The Olympic athlete data set \cite{griffin2018120} contains 271,116 samples (individual athletes) with 15 features. This data set has effective feature complexity and non-linearity, making it a suitable proof of concept data set \cite{griffin2018120}. Our second data set uses US adult census income data \cite{becker1998visualizing}, which contains features on age, gender, occupation, education, etc.

\subsection{Model}
Generative Adversarial Networks (GANs) are the model we use to generate synthetic demographic data in this study. A GAN is a game-theoretic-based generative model that includes generator and discriminator networks \cite{goodfellow_generative_2020}. The generator takes a noise variable as input and generates synthetic samples by matching distributions with the actual data. The discriminator, trained by real data, distinguishes the actual and synthetic samples produced by the generator \cite{goodfellow_generative_2020}. The strong ability of GANs to learn data distributions makes them a promising candidate for demographic data generation.

\subsection{Evaluation Metrics}
We used an open-source tool SDMetrics to measure model performance \cite{sdmetrics}. Four metrics are used in this research: KSComplement, TVComplement, CorrelationSimilarity and ContingencySimilarity. The first two metrics compute the similarity between a real column and a synthetic column in terms of the column shapes (shape score), which is the marginal distribution or 1D histogram of the column \cite{sdmetrics}. The latter two compute the similarity of a pair of columns between the real and synthetic datasets (pair trend score), which are the 2D distributions \cite{sdmetrics}. KSComplement and CorrelationSimilarity were designed for continuous columns, while TVComplement and ContigencySimilarity were designed for categorical columns. All four metrics have score range from 0 to 1. A zero value indicates maximum dissimilarity between the real and synthetic data, while a value of one indicates perfect match between the real and synthetic data. 

\section{Experiment and results}
\label{sec:others}
In this section, we will first introduce the data processing and model building procedures and then the evaluation and visualization of the results.
\subsection{Data preparation}
For the Olympics data set, we focus on features that ensure complexity. We use three continuous (age, height, weight) and six categorical (sex, year, season, city, sport, medal) features as the input data. Missing values in the columns of age, height, and weight were filled by the median value of athletes from the corresponding sport with the same gender. Besides gold, silver, and bronze, the missing values in the column of medal were filled by the string ``Thanks'', which represents no medal won by the corresponding athlete. In addition, an athlete may participate in several different sports and events in the same Olympic game in the same year. In order to keep complexity, we dropped duplicate representation of athletes, and added new features, AOS and AOE, to count the number of sports and events in which they participated. Athletes that participated in different games in different years were kept to ensure the richness of the dataset. After all the processes above, the Olympic dataset contained 188,169 rows and 11 columns (three continuous and eight categorical columns). 

\subsection{Data pre-processing}
Normalization was applied to the continuous columns, and one hot encoding was used on the categorical columns. We used three normalization functions: max-absolute, min-max and standardization \cite{grus2019data}.

Since all features in the categorical columns are nominal without an order of importance, we implemented one-hot encoding on categorical columns. The input data is encoded before training, and synthetic data is decoded after being generated by the generator.

\subsection{GAN structure}
In this research, we use fully-connected networks in the generator and discriminator to capture all possible correlations between columns because tabular data do not have complicated local structures \cite{xu2019modeling}. Specifically, there is a fully-connected hidden layer, followed by a leaky ReLU activation function, and then another fully-connected hidden layer. After the last fully-connected hidden layer in the discriminator, a Sigmoid function is used to represent the authenticity of generated synthetic rows.

According to the parameter turning by grid search, the negative slope value for the leaky ReLU activation function is set to 0.8 in the generator and 0.1 in the discriminator. The loss function we used for both the generator and discriminator is cross-entropy \cite{zhang2018generalized}. The optimizer we used is Adam \cite{zhang2018improved}, with a learning rate of 0.0001.

\subsection{Train with generation by geometric progression}
\begin{figure}[htp]
    \centering
    \begin{minipage}{0.53\textwidth}
        \centering
        \includegraphics[width=0.98\textwidth]{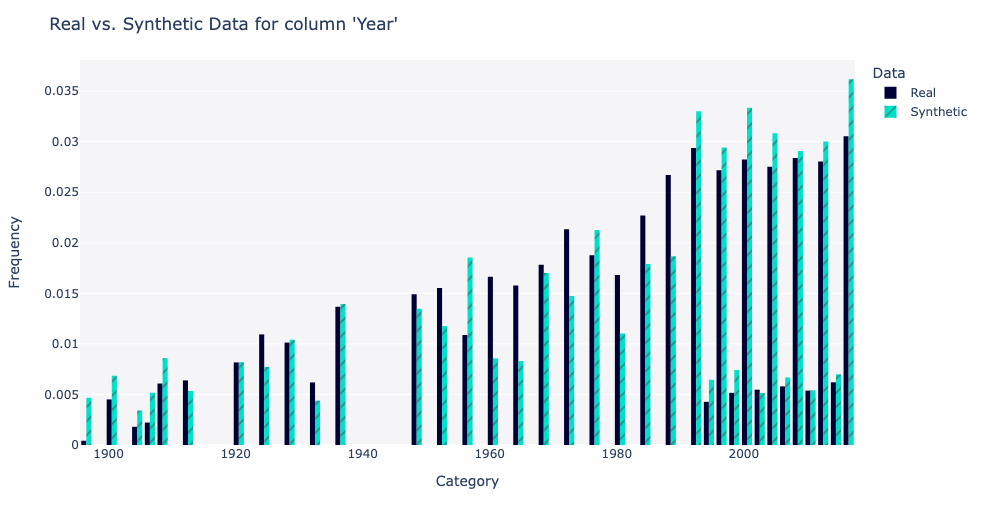} 
        \caption{Feature ``Year''.}
        \label{fig:year1}%
    \end{minipage}\hfill
    \begin{minipage}{0.46\textwidth}
        \centering
        \includegraphics[width=0.95\textwidth]{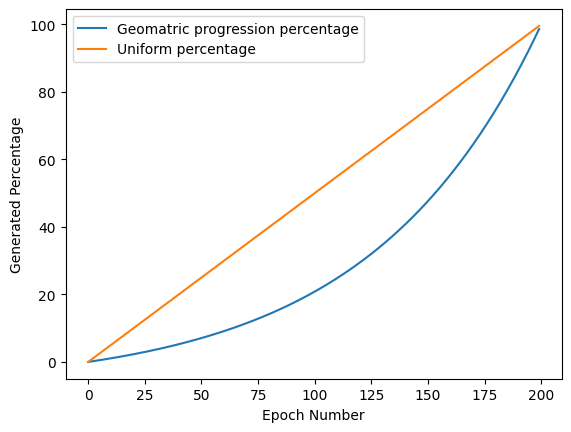} 
        \caption{Geometric progression as generation percentage.}
        \label{fig:geometric_progression}
    \end{minipage}
\end{figure}

We evaluated and visualized the generation results of the initial GANs model.
The average shape and pair trend score for continuous columns can reach approximately 0.90 and 0.98, respectively. However, for categorical columns, the average shape and pair trend scores were only 0.53 and 0.73, respectively. For the columns with imbalanced data, such as sex, season, medal, AOS and AOE, we can observe that the trained generator can only produce the most salient category -- minor categories are not generated. The performance is acceptable for the balanced column ``year'' as shown in Figure \ref{fig:year1}. For another balanced column ``sport'', every category was generated, but we can still observe that more data was generated from salient categories instead of minor ones. All visualizations can be found on our GitHub website.

Imbalance problem is one of the most important problems in tabular synthetic data generation using GANs. To solve this problem, \cite{xu2019modeling} applied a method called training-by-sampling to ensure that all the categories from discrete attributes are sampled evenly during the training process, which means that the discriminator assessed each generation produced by the conditional generator. In other word, this research evenly explored all possible values in discrete columns during the training process. In our research, we implemented a similar method which we named \textit{train with generation}. The main idea of this method is to train and generate simultaneously, evenly generating all discrete attributes during the early training process before the model was well-trained. Specifically, a certain percentage of data was generated after each training epoch. For example, if we have 50 training epochs, 2\% synthetic data will be generated after each training epoch.

After applying the train with generation method, the imbalance problem was significantly improved. However, to further control the generation amount after each training epoch, we also applied a geometric progression as the percentage of generated synthetic data. We can generate fewer random samples during the forward training process and more well-trained samples during the backward training process. 



For example, if the number of training epochs is 200, the percentage of synthetic data is 100\%, the first generation percentage is 0.1\%, then the common ratio should be approximately 1.01344. The visualization of this example is shown in Figure \ref{fig:geometric_progression}.


A simple experiment was designed to determine this dataset's best geometric progression parameter. In this experiment, three epoch numbers and four first-item values were included, as shown in the Table \ref{tab:correlation1}. According to the experiment results, we choose the parameters for this dataset: epoch number is 50, first item value is 0.2, and the common ratio is 1.15884.

\begin{figure}[htp] 
    \centering
    \subfloat[Feature "Age".]{%
        \includegraphics[width=0.5\textwidth]{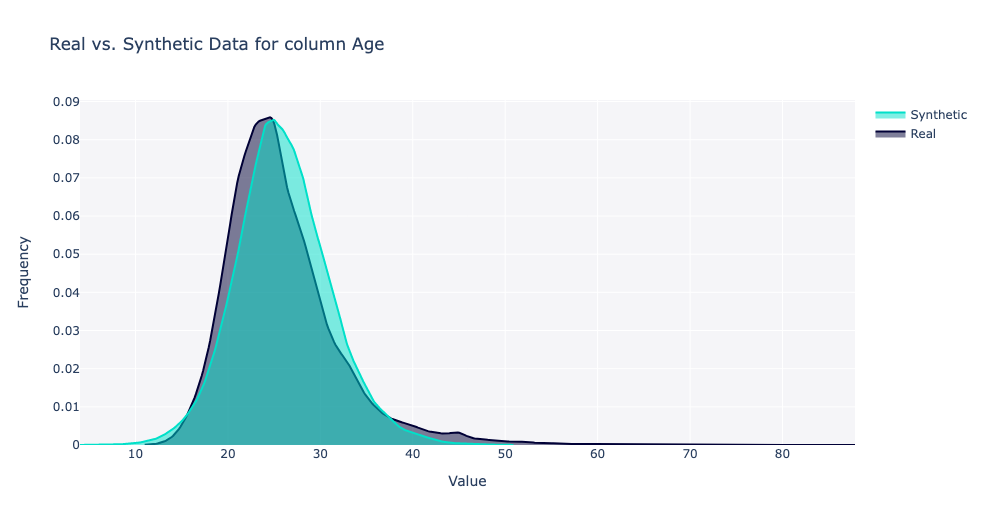}%
        \label{fig:age2}%
        }%
    \subfloat[Feature "Season".]{%
        \includegraphics[width=0.5\textwidth]{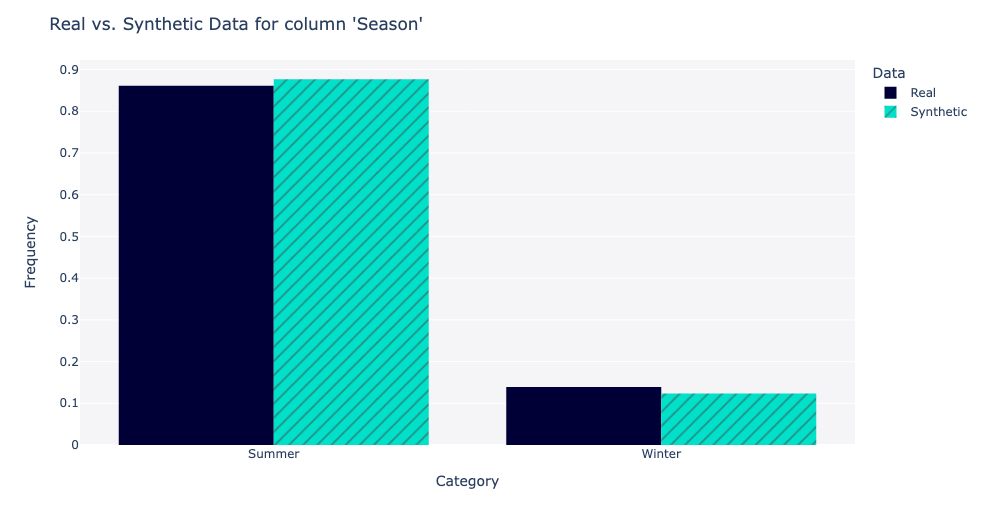}%
        \label{fig:season2}%
        }%
        \hfill%
    \subfloat[Feature "Sport".]{%
        \includegraphics[width=0.5\textwidth]{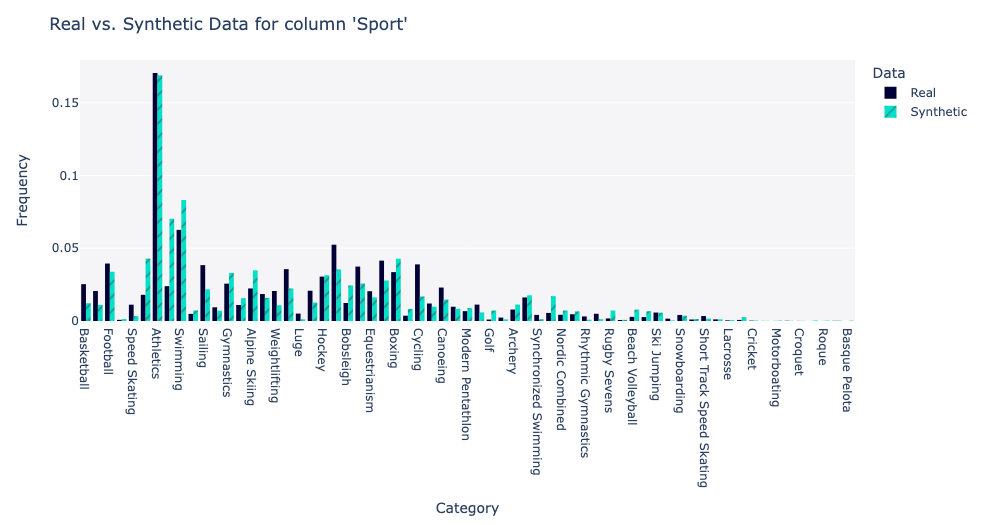}%
        \label{fig:sport2}%
        }%
    \subfloat[Correlation similarity heat-map]{%
        \includegraphics[width=0.5\textwidth]{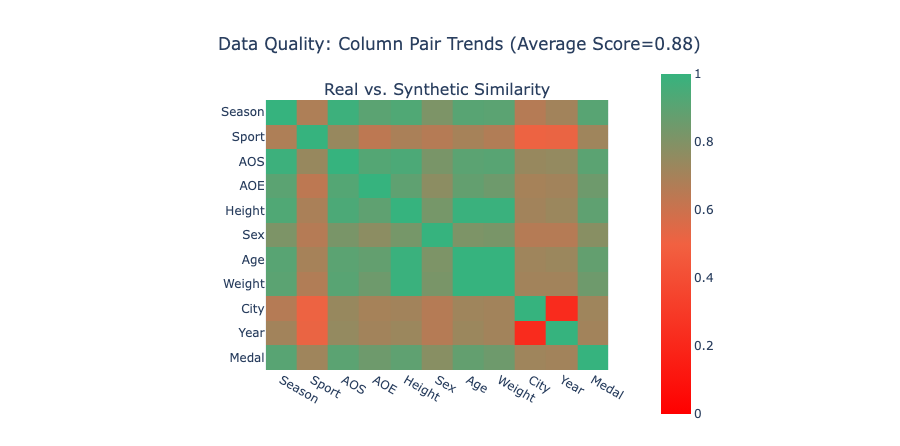}%
        \label{fig:heatmap}%
        }%
    \caption{Visualization using train with generation by geometric progression method}
    \label{fig:result}
\end{figure}

After implementing this train with generation by geometric progression method, we implemented another evaluation and visualization, and the results are shown in Figure \ref{fig:result}. We can observe that each minor category can be generated in the imbalanced columns. The average shape and pair trend scores for categorical columns were increased to 0.78 and 0.88, respectively. In addition, the performance for continuous data was kept similar to the last implementation. Overall, the average shape and pair trend score for all columns reached 0.88 and 0.92, respectively.




The column correlation similarity between real and synthetic data amount in all columns are shown in Figure \ref{fig:heatmap}. 
To better understand this heat map, we defined three column types:
\begin{itemize}
    \item Type A: continuous column (age, height, weight)
    \item Type B: categorical column with small categories (sex, medal, AOE, season)
    \item Type C: categorical column with large categories (year, sport, city)
\end{itemize}


\begin{table}
  \begin{minipage}{.5\linewidth}
    \centering
    \begin{tabular}{ |c|c|c|c|c| }
    \toprule
    \diagbox{Epochs}{Values} & \textbf{0.1}  & \textbf{0.2} & \textbf{0.3} & \textbf{0.4}\\
     \midrule
     \textbf{50} & 0.873 & 0.889 & 0.872 & 0.873\\ 
     \textbf{100} & 0.870 & 0.885 & 0.871 & 0.865 \\  
     \textbf{200} & 0.859 & 0.878 & 0.862 & 0.861 \\
     \bottomrule
    \end{tabular}
    \caption{geometric progression parameter turning}\label{tab:correlation1}
  \end{minipage}%
  \begin{minipage}{.5\linewidth}
    \centering
    \begin{tabular}{ |c|c|c| }
    \toprule
     \textbf{Types} & \textbf{Correlation Similarity} \\
     \midrule
     Type AA & 0.99 \\ 
     Type AB & 0.85 \\  
     Type AC & 0.75 \\
     Type BB & 0.94 \\
     Type BC & 0.80 \\
     Type CC & 0.55 \\
     \bottomrule
    \end{tabular}
    \caption{Average column correlation similarity between three type columns}\label{tab:correlation2}
  \end{minipage}
\end{table}
We calculated the average column correlation similarity between these three types and obtained the results below.
From the above table, the correlation performance among continuous columns is excellent. However, the correlation between continuous and categorical columns depends on the number of categories. The correlation performance related to columns with large categories (Type C) may be lower than other types. 

\subsection{Evaluation and visualization with new dataset}
\begin{figure}[htp] 
    \centering
    \subfloat[Feature "capital-gain".]{%
        \includegraphics[width=0.5\textwidth]{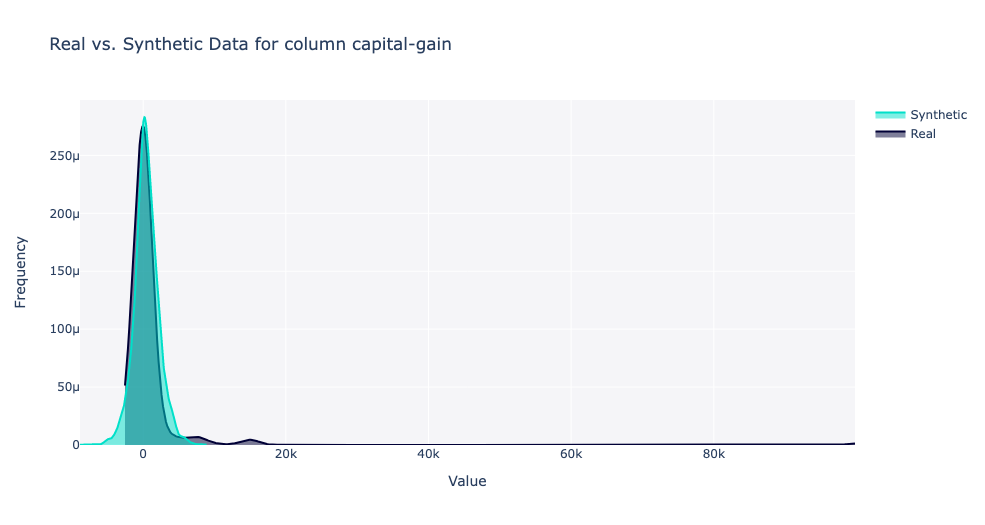}%
        \label{fig:capitalGain}%
        }%
    \subfloat[Feature "workclass".]{%
        \includegraphics[width=0.5\textwidth]{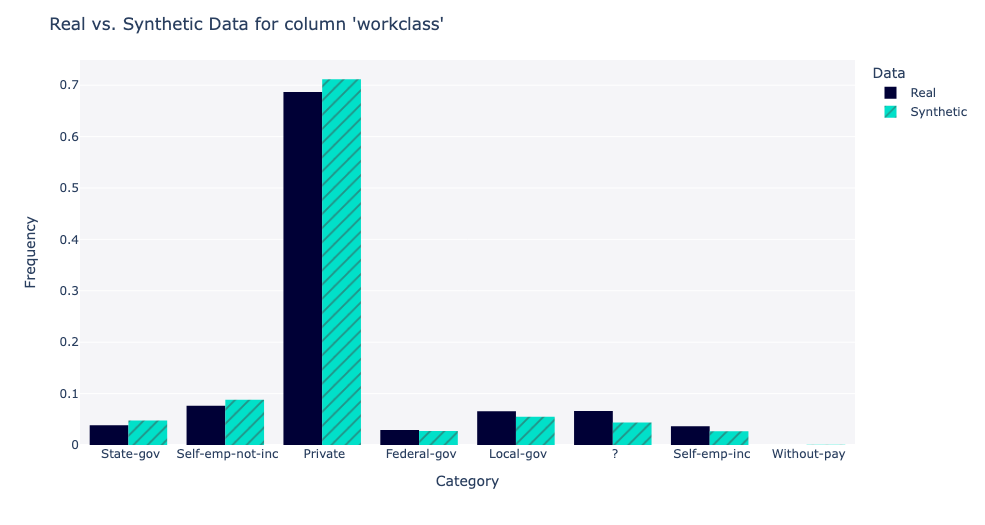}%
        \label{fig:workclass}%
        }%
    \caption{Visualization of generation with new dataset}
    \label{fig:newdata}
\end{figure}

The model was further tested with the US adult census income dataset, which includes information about age, education, occupation, income, etc. The average shape and pair trend score for all columns in this new dataset reached 0.88 and 0.90, respectively, which is the proof of the performance of this model. Figure \ref{fig:newdata} shows the distributions of two representative features.



\section{Discussion}
From the initial generation results, we found that the severe mode collapse happened for the imbalanced categorical columns, such as ``Season'' and ``Sport''. Highly imbalanced categorical data is ubiquitous in real-world datasets. In the dataset we used, 7/12 of the categorical columns are imbalanced, and three of them are highly imbalanced -- the major category appears in more than 80\% of the rows. There are two conditions when mode collapse happens for imbalanced categorical columns. The first is that the generator only generated synthetic data for salient categories instead of minor types. For example, in the season column, only the ``Summer'' category was generated rather than the ``Winter'' category. The second condition is that the salient category was generated much more than the real size compared to other categories.

Generally, a GAN is fed with a random vector from the standard multivariate normal distribution $\mathcal{N}(0,1)$, and by means of the Generator and Discriminator neural networks, one eventually obtains a deterministic transformation that maps the normal distribution onto the distribution of the data \cite{goodfellow_generative_2020}. In this way, the imbalance of categorical data was not considered when training the generator.

To solve the imbalance problem, we applied the train with generation method, which generates random categories in the early period of the training process. Before the model is fully trained, minor categories are generated with higher probability. After the early training process, the ratio of minor to major category generation is decreased, such that the final amount of samples in different categories is balanced towards the real data. In addition, by using geometric progression, over-generating random samples is avoided and the balance between minor and major categories is precisely controlled.

\section{Conclusion}
We built a Generative Adversarial Network to generate high-quality synthetic demographic data. We obtained high model performance for continuous columns, and by using the train-with-generation method with a geometric progression, obtained significantly improved generation performance for categorical columns.

\bibliographystyle{unsrt}  
\bibliography{references}  






\end{document}